\documentclass[journal]{IEEEtran}
\usepackage{amsmath,amsfonts}
\usepackage{algorithmic}
\usepackage{array}
\usepackage[caption=false,font=normalsize,labelfont=sf,textfont=sf]{subfig}
\usepackage{textcomp}
\usepackage{stfloats}
\usepackage{url}
\usepackage{verbatim}
\usepackage{graphicx}
\usepackage{color}
\usepackage{booktabs}
\usepackage{multirow}
\usepackage[numbers]{natbib}
\usepackage{algorithm}
\usepackage{algorithmic,amsmath,bm,amssymb}
\def\BibTeX{{\rm B\kern-.05em{\sc i\kern-.025em b}\kern-.08em
    T\kern-.1667em\lower.7ex\hbox{E}\kern-.125emX}}
\usepackage{balance}
\begin{document}
\title{MIMIC: Mask Image Pre-training with Mix Contrastive Fine-tuning for Facial Expression Recognition}

\author{Fan Zhang, Xiaobao Guo,  Xiaojiang Peng,~\IEEEmembership{Member,~IEEE,} Alex Kot,~\IEEEmembership{Fellow,~IEEE}
        
\IEEEcompsocitemizethanks{
\IEEEcompsocthanksitem This work is done when Fan Zhang is an intern at Shenzhen Technology University. Fan Zhang is with the Department of Electrical and Computer Engineering, Georgia Institute of Technology, Shenzhen, China. E-mail: fanzhang@gatech.edu
\IEEEcompsocthanksitem X. Guo and A. Kot are with Rapid-Rich Object Search (ROSE) Lab, Nanyang Technological University, Singapore. E-mail: \{xiaobao001, eackot\}@ntu.edu.sg
\IEEEcompsocthanksitem Xiaojiang Peng (Corresponding Author) is with the College of Big Data and Internet, Shenzhen Technology University, Shenzhen, China. E-mail: pengxiaojiang@sztu.edu.cn

}
\thanks{Manuscript created January 2024.}
}

\maketitle

\begin{abstract}
Cutting-edge research in facial expression recognition (FER) currently favors the utilization of convolutional neural networks (CNNs) backbone which is supervisedly pre-trained on face recognition datasets for feature extraction. However, due to the vast scale of face recognition datasets and the high cost associated with collecting facial labels, this pre-training paradigm incurs significant expenses. Towards this end, we propose to pre-train vision Transformers (ViTs) through a self-supervised approach on a mid-scale general image dataset. In addition, when compared with the domain disparity existing between face datasets and FER datasets, the divergence between general datasets and FER datasets is more pronounced. Therefore, we propose a contrastive fine-tuning approach to effectively mitigate this domain disparity. Specifically, we introduce a novel FER training paradigm named \underline{M}ask \underline{I}mage pre-training with \underline{MI}x \underline{C}ontrastive fine-tuning (MIMIC). In the initial phase, we pre-train the ViT via masked image reconstruction on general images. Subsequently, in the fine-tuning stage, we introduce a mix-supervised contrastive learning process, which enhances the model with a more extensive range of positive samples by the mixing strategy. Through extensive experiments conducted on three benchmark datasets, we demonstrate that our MIMIC outperforms the previous training paradigm, showing its capability to learn better representations. Remarkably, the results indicate that the vanilla ViT can achieve impressive performance without the need for intricate, auxiliary-designed modules. Moreover, when scaling up the model size, MIMIC exhibits no performance saturation and is superior to the current state-of-the-art methods. 
\end{abstract}


\section{Introduction}
\label{intro}
\IEEEPARstart{T}{he} expression of emotions through facial expressions represents a fundamental means of communication for human beings. Automatic emotion recognition plays a great role in various applications including surveillance\cite{clavel2008fear}, educations \cite{yang2018emotion}, medical treatments \cite{pioggia2005android} and marketing \cite{ren2012linguistic}. In the realm of the sub-areas, facial expression recognition (FER) is a very challenging task, particularly in uncontrolled environments. Even worse, the considerable similarities of inter-class emotion samples present a unique challenge to effectively distinguish between facial expressions.

The recent achievements in FER~\cite{li2020deep,li2017reliable,li2021adaptively,she2021dive,xue2021transfer,tao2023hierarchical,gao2023ssa,maronidis2011improving,matsugu2003subject,ioannou2005emotion} have greatly benefited from convolutional neural networks (CNNs) backbone that pre-trained on face recognition datasets in a supervised manner. 
The success of this pre-training paradigm hinges on two essential conditions. 
On one hand, the face dataset used for pre-training needs to be massive in scale, exemplified by datasets like MS-Celeb-1M\cite{guo2016ms}, consisting of 5 million images. 
On the other hand, it necessitates the collection of labels for this extensive dataset to provide supervisory signals during pre-training.
Evidently, this pre-training paradigm incurs substantial costs.

Motivated by the recent success of self-supervised learning in image classification tasks \cite{he2022masked,xie2022simmim,bao2021beit,dong2021peco}, we aim to reduce the cost of pre-training for FER.
Certainly, we propose two enhancements to the pre-training approach: firstly, substituting large-scale face datasets (e.g., MS-Celeb-1M with 5 million images) with mid-scale general image datasets (e.g., ImageNet-1K with 1 million images) for pre-training. Secondly, adopting a self-supervised pre-training method based on masked image reconstruction with vision Transformers (ViTs) instead of supervised pre-training approaches to mitigate the expenses associated with label collection.

\begin{figure}[t]
    \centering
    \includegraphics[width=\linewidth]{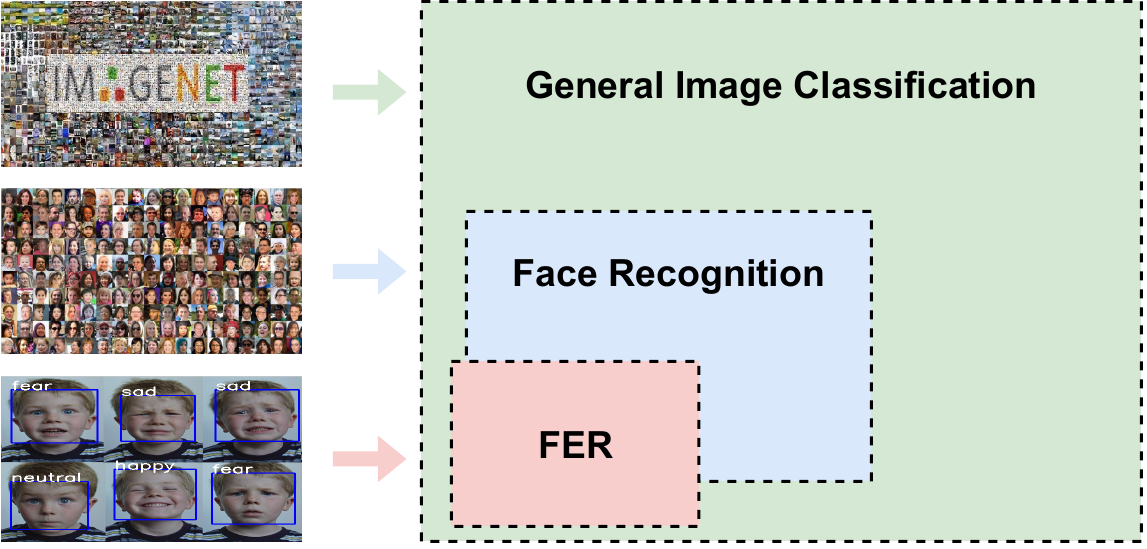}
    \caption{The relationship of three domains. There exists a higher domain disparity between FER and general image classification. }
    \label{domain}
\end{figure}

However, due to the fact that face datasets and FER datasets primarily consist of facial images, while general image datasets encompass a broader range of generic objects, the domain disparity between general image datasets and FER datasets is more substantial than that between face datasets and FER datasets (as shown in Figure~\ref{domain}).
Therefore, naively fine-tuning the network on FER datasets through the classification loss may lead to sub-optimal solutions, as the common sense that a good generalization performance requires a stronger discriminative ability between classes in downstream tasks~\cite{DBLP:conf/iclr/GunelDCS21,DBLP:conf/icassp/MoukafihGS23}.
Thus, the main challenge is to bridge the domain gap of pre-training on general images and fine-tuning on FER datasets.

To address the challenge, we adopt the powerful technique of contrastive learning during the fine-tuning stage. Traditional self-supervised contrastive learning methods \cite{wu2018unsupervised,oord2018representation,hjelm2018learning,ye2019unsupervised,he2020momentum,chen2020improved} focus on pulling the positive pairs (the sample with its augmentation) closer and pushing the negative pairs (the sample with all other different samples) away from each other in the representation space. 
This approach is bound to incorrectly consider samples from the same category as negative samples to the anchor, leading to an increased distance in the feature space.
To this end, we resort to Supervised Contrastive Learning (SCL) \cite{khosla2020supervised} which attempts to maximize the similarity among intra-class samples and minimize the similarity among inter-class samples. 
By adopting SCL, we effectively tackle the issue of negative sample selection.
Furthermore, in the context of FER, negative samples from different expressions may share high similarities. 
For example, the expressions 'happiness' and 'neutral' may be considered as negative pairs, while they are particularly similar and often exhibit subtle differences in specific regions.
Even the naked eyes find it challenging to differentiate between them.

Therefore, we improve SCL by a mixing strategy so that facial expression samples from different classes that share high similarities could be viewed as positive pairs.
Specifically, we introduce an extra mix-supervised contrastive branch in the fine-tuning stage.
The deep features from mixed samples are projected into a new hidden space for contrastive learning.
The proposed mix-supervised contrastive fine-tuning strategy adeptly considers highly similar yet different-class expressions as positive samples. These techniques generate a vast number of positive and negative samples according to predefined rules, enabling the model to learn discriminative features for FER.

In summary, this paper proposes a novel FER training paradigm termed \underline{MI}x \underline{C}ontrastive fine-tuning (MIMIC) to achieve effective FER without bells and whistles.
The process of MIMIC includes two steps: pre-training a ViT encoder on a general image dataset with masked image reconstruction, and then fine-tuning the network on downstream FER datasets with an extra mix-contrastive branch to simultaneously recognize facial expressions and mitigate the domain disparity. The main contributions of this paper are as follows:
\begin{itemize}
    \item Our investigation reveals that self-supervised ViT with an appropriate pre-training pipeline can yield better performance than pre-trained CNN models for FER. We reduce the pre-training costs by adopting a self-supervised approach with masked image reconstruction on a mid-scale general image dataset, demonstrating that the success of FER no longer relies on large-scale face datasets with annotated labels.
    \item Considering the limited number of categories and notable similarities within each class in FER, we carefully develop a mix-supervised contrastive fine-tuning strategy that facilitates the effective transfer of pre-trained models to FER. This strategy is specifically tailored to address the unique challenges posed by FER, allowing for improved performance in capturing the nuances among facial expressions. 
    \item We conduct extensive experiments on three popular datasets, and demonstrate that the vanilla ViT-B/16 model can achieve remarkable performance without the need for additional parameters or complex feature extraction structures. Furthermore, we observe that scaling up the model size results in no performance saturation, indicating the potential for further improvements. By utilizing ViT-L/16 as the backbone, our method achieves outstanding results, with a performance of $91.26\%$ on the RAF-DB dataset and $91.24\%$ on the FERPlus dataset, which surpasses several state-of-the-art (SOTA) approaches, highlighting the superiority of our approach in FER.
\end{itemize}

\section{Related Work}
\label{related work}

\subsection{Facial Expression Recognition}
In general, a FER system typically comprises three stages: face detection, feature extraction, and expression recognition. In the face detection stage, various face detectors, such as MTCNN \cite{zhang2016joint} and Dlib \cite{amos2016openface}, are employed to locate faces in complex scenes.

For the feature extraction, many approaches \cite{li2017reliable,li2021adaptively,she2021dive,xue2021transfer,tao2023hierarchical,gao2023ssa,maronidis2011improving,matsugu2003subject,ioannou2005emotion} have been proposed, which can be separated into two types: handcrafted features and learning-based features. 
Regarding the handcrafted features, they can be categorized into texture-based local features, geometry-based global features, and hybrid features. Texture-based features encompass SIFT \cite{ng2003sift}, HOG \cite{dalal2005histograms}, Histograms of LBP \cite{shan2009facial}, Gabor wavelet coefficients \cite{liu2002gabor}, among others. Geometry-based features primarily rely on landmark points around the nose, eyes, and mouth. Combining two or more of these engineered features constitutes hybrid feature extraction, contributing to a more enriched representation.
Traditional studies based on handcrafted features \cite{hu2008multi,luo2013facial,pietikainen2011computer} focus on texture information from in-the-lab datasets such as CK+ \cite{lucey2010extended} and Oulu-CASIA \cite{zhao2011facial}. However, with the emergence of large-scale unconstrained FER datasets \cite{barsoum2016training,li2017reliable,mollahosseini2017affectnet}, deep facial expression recognition (DFER) algorithms have been developed to design effective CNNs to extract learning-based features, thereby achieving superior performance. For example, Li et al. \cite{li2017reliable} propose a locality-preserving loss to learn more discriminative facial expression features, while Wang et al. \cite{wang2020region} propose a region-based attention network to capture important facial regions. Li et al. \cite{li2018occlusion} also explore partially occluded facial expression recognition. In addition, several works \cite{she2021dive,wang2020suppressing,zeng2018facial} address the inconsistent annotation problem in DFER. Recently, Xue et al. \cite{xue2021transfer} explore relation-aware representations for Transformers-based DFER.
Following the feature extraction stage, the subsequent step involves inputting the features into a supervised classifier, such as Support Vector Machines (SVMs), a softmax layer, or logistic regression, to assign the samples with category-level expression labels.
\begin{figure*}[t]
    \centering
    \includegraphics[width=\linewidth]{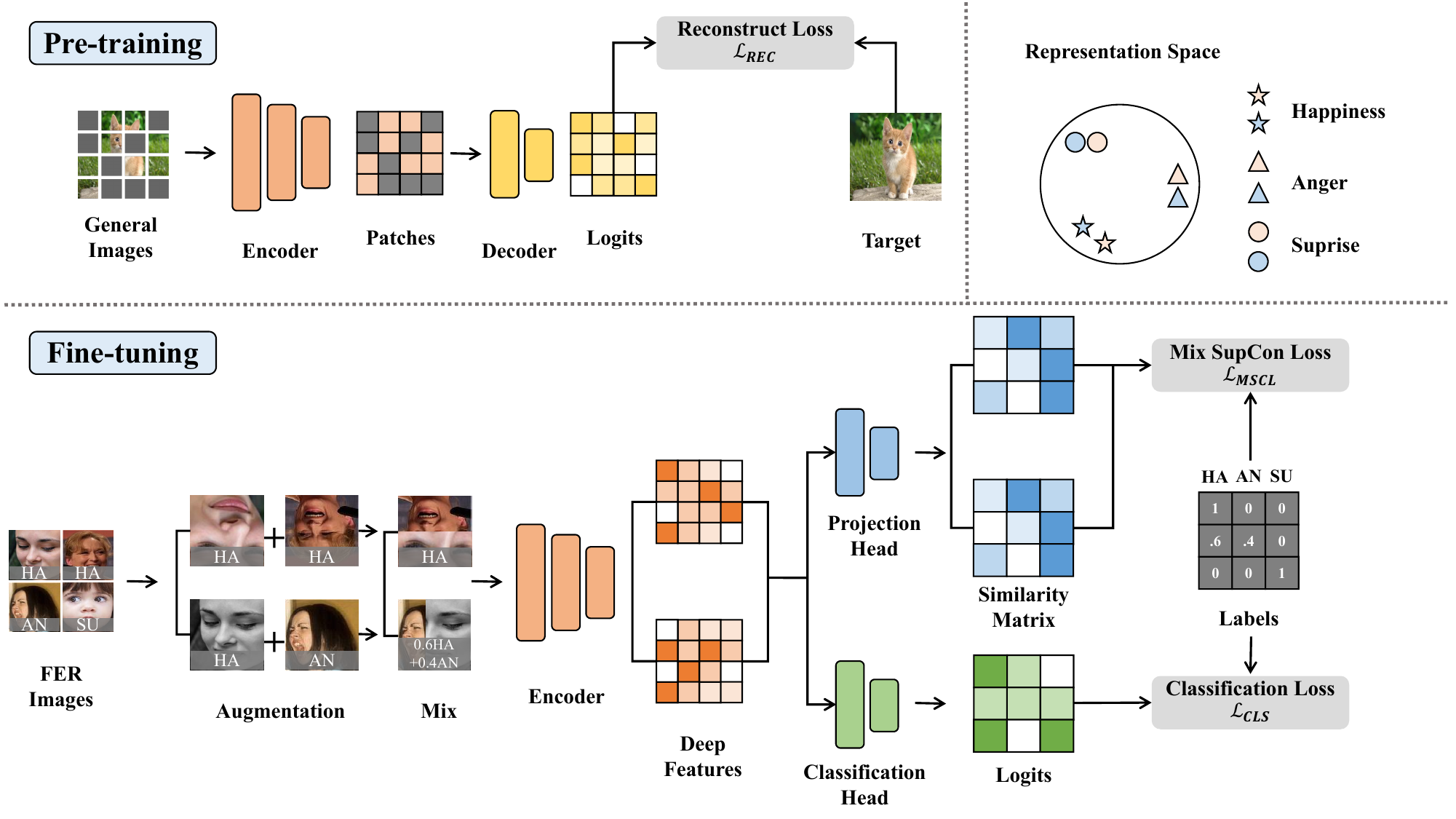}
    \caption{Illustration of our pipeline. The encoder is firstly pre-trained by mask image modeling in other domains (\textit{e.g.,} ImageNet). Then we mix the augmented FER images and send them to the encoder. After the encoder, a projection head and a classification head are utilized to simultaneously refine visual representations learned from other domains and recognize facial expressions.}
    \label{method}
\end{figure*}

Existing DFER algorithms usually train a network to recognize facial expressions in an end-to-end manner. The performance of the network is poor when training from scratch. Thus, almost all DFER methods initialize the network with weights pre-trained on large datasets from other domains. However, the weights are trained in a fully-supervised manner (\textit{e.g.}, image classification and face recognition). We, for the first time, attempt to exploit the power of self-supervised pre-training from other domains and adapt it to enhance FER performance.

\subsection{Self-supervised Learning}
The two main approaches to self-supervised representation learning are contrastive learning and mask image modeling. The core idea of contrastive learning \cite{hadsell2006dimensionality} is to attract the positive sample pairs and push away the negative sample pairs. In recent years, contrastive learning has emerged as a powerful technique for unsupervised representation learning \cite{wu2018unsupervised,oord2018representation,hjelm2018learning,ye2019unsupervised,he2020momentum,chen2020improved}. 
In practical scenarios, contrastive learning techniques gain advantages from a substantial quantity of negative samples \cite{wu2018unsupervised,oord2018representation,hjelm2018learning,ye2019unsupervised}. These samples can be stored in a memory bank \cite{wu2018unsupervised}. MoCo \cite{he2020momentum}, within a Siamese network, manages a queue of negative samples and transforms one branch into a momentum encoder to enhance the consistency of the queue. SimCLR \cite{chen2020simple} directly utilizes negative samples present in the current batch, and it necessitates a large batch size for optimal performance.
Supervised contrastive learning \cite{khosla2020supervised} considers intra-class samples as positive pairs and inter-class samples as negative pairs, which reinforce learning representations.

As vision transformers have advanced \cite{dosovitskiy2020image,liu2021swin}, masked image modeling is progressively supplanting the predominant role of contrastive learning in self-supervised visual representation learning. This shift is attributed to its superior fine-tuning performance across diverse visual downstream tasks. Mask image modeling obtains good learning representations from predicting the mask target signals of the original image. Different pretext tasks design different target signals, such as normalized pixels \cite{he2022masked,xie2022simmim}, discrete tokens \cite{bao2021beit,dong2021peco}, HOG feature \cite{wei2022masked}, deep features \cite{baevski2022data2vec,zhou2021ibot}
or frequencies \cite{liu2022devil,xie2022masked}. 

For FER, the mainstream approaches focus on designing complex network architectures instead of leveraging powerful training strategies. We propose to explore transferring mask image modeling and contrastive learning from general image classification to FER without introducing additional parameters. 

\section{Method}
\label{sec:method}

The pipeline of our proposed MIMIC is shown in Figure \ref{method}. In the following two sections, a comprehensive analysis will be presented to elucidate the particular dissimilarities between the conventional DFER pipeline and MIMIC, with respect to the pre-training and fine-tuning stages.
\begin{figure*}[t]
    \centering
    \includegraphics[width=0.8\linewidth]{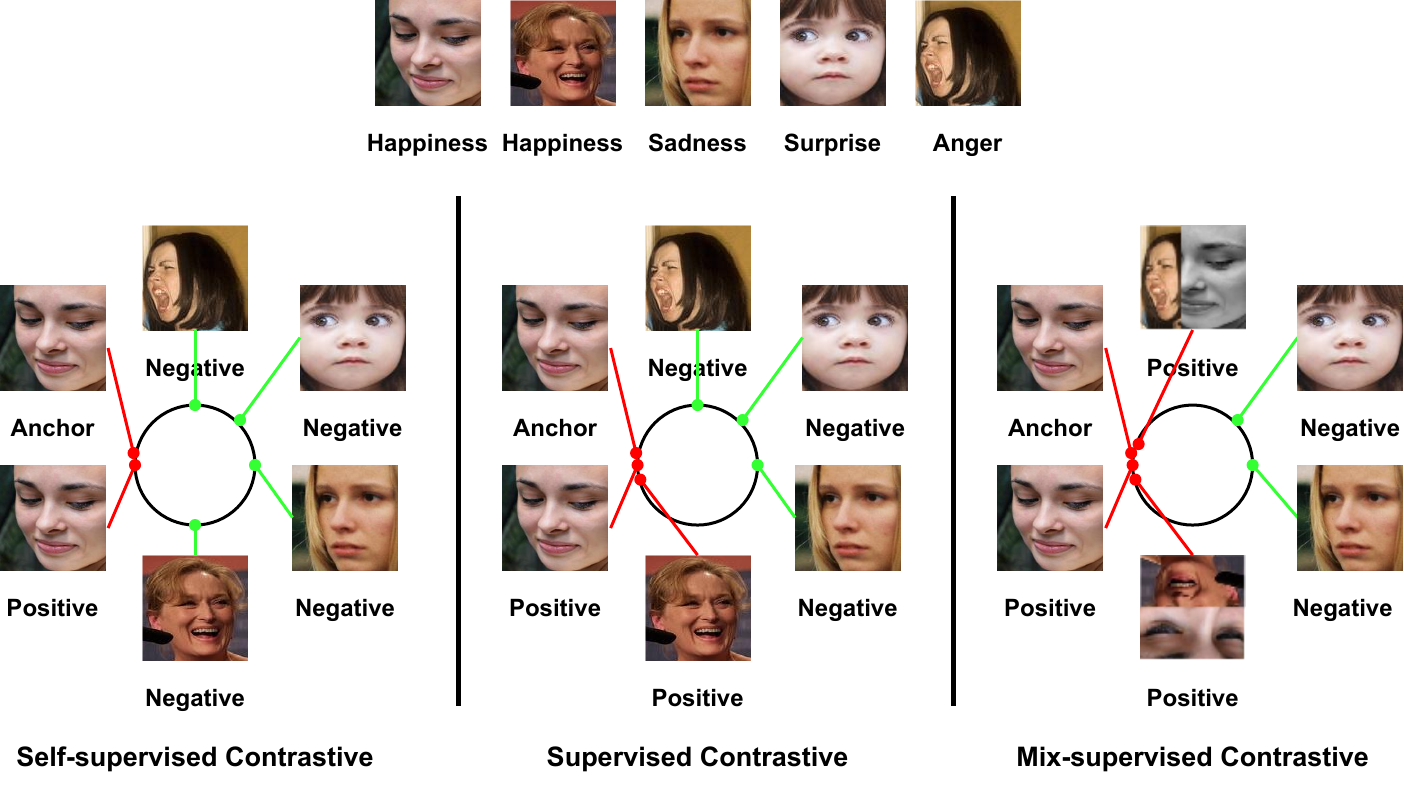}
    \caption{Comparison of three formats of contrastive learning. Self-supervised format ignores intra-class similarity, while supervised contrastive format leverages class labels to attract intra-class samples. However, the supervised contrastive format only focuses on intra-class similarity and ignores inter-class similarity. Mix-supervised contrastive format selects diverse positive samples, for inter-class samples with high similarities may boost visual representations as well.}
    \label{cl}
\end{figure*}

\subsection{Mask Image Pre-training for FER}
Training from scratch limits the performance of most DFER methods, for the reason that FER datasets are usually too small and the backbone is not fully exploited. The success of previous DFER algorithms owing to pre-trained weights on face recognition datasets. In traditional supervised pre-training, a set of instance-label pairs can be formulated as $\mathcal{C = (X, Y)}$, where $\mathcal{X}=\{x_i\}^{N}_{i=1}$ and $\mathcal{Y}=\{y_i \in \{0,1\}^C\}^{N}_{i=1}$ are the set of training samples and the corresponding $\mathcal{C}$-class labels in on-hot format. The parameter $\theta$ of the network is updated by classification loss:
\begin{equation}
    \mathcal{L}_{cls}^{p}=-\frac{1}{N}\sum_{i=1}^{N}\sum_{c=1}^{C}y_{i}^{c}log(p_{c}(x_{i},\theta)) .
\end{equation}
Supervised pre-training methods aim to pre-train a powerful classifier to recognize general images. The network thus learns good representations and generalizes well to downstream tasks after pre-training. However, the supervision signal is just the category label of each image. The pretext task is not difficult enough for the network. Therefore, the potential of neural networks has yet to be fully exploited.

To address the problem, recently a new paradigm of self-supervised pre-training called mask image modeling (MIM) has been raised. MIM designs a more difficult pretext task for the network, \textit{i.e.}, reconstructing the original image from masked signals. The masked signals are various in different formats. They can be pixels, discrete tokens, hand-crafted features, deep features, or frequencies. In this way, the network is updated by reconstructing loss:
\begin{equation}
    \mathcal{L}_{rec}^{p}=-\frac{1}{N}\sum_{i=1}^{N}\sum_{d=1}^{D}(x_{i}^{d}-p_{d}(x_{i},\theta))^2 ,
    \label{lrec}
\end{equation}
where $D$ denotes the number of patches or discrete tokens of an image. Predicting masked parts of the image is rather fine-grained than classification. The network thus tries its best to learn better representations for the purpose of reconstructing the raw signals. What's more, there is no need to offer human-annotated labels for training. Thus we can leverage a large number of unlabeled images from other domains for training and transfer the knowledge to FER.

\begin{figure*}[t]
    \centering
    \includegraphics[width=\linewidth]{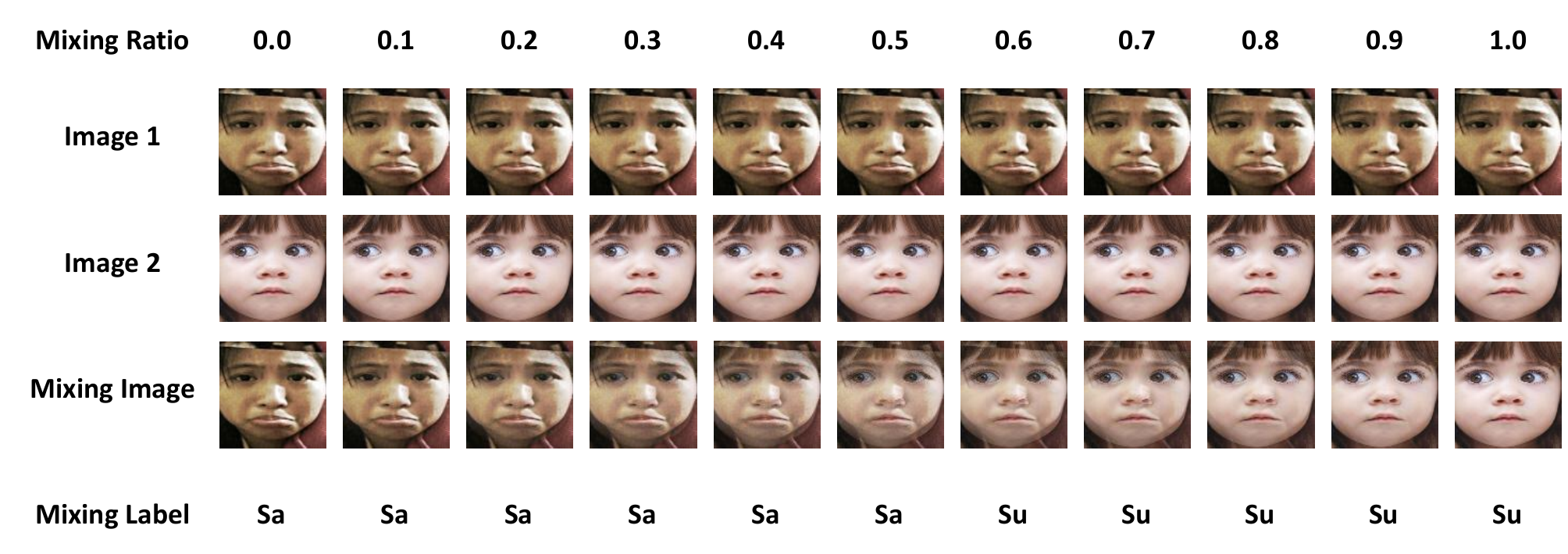}
    \caption{An illustration of our mixing strategy. Sa and Su denotes Sadness and Surprise, respectively.}
    \label{fig:vis_exp}
\end{figure*}

\subsection{Contrastive Fine-tuning for FER}
After pre-training, only the backbone of the network is taken out for fine-tuning and the remaining parts are dropped. Traditional DFER algorithms design explicit deep blocks and connect them with the backbone. The final block of the network is a classification head which is used to recognize facial expressions. Usually, DFER methods perform an end-to-end naive fine-tuning with another classification loss:
\begin{equation}
    \mathcal{L}_{cls}^{f}=-\frac{1}{N}\sum_{i=1}^{N}\sum_{k=1}^{K}y_{i}^{k}log(p_{k}(x_{i},\theta)) .
    \label{lfcls}
\end{equation}
The difference between $\mathcal{L}_{cls}^{f}$ and $\mathcal{L}_{cls}^{p}$ is the number of categories, for the pre-train domain is usually larger and contains more categories. However, the naive fine-tuning strategy merely utilizes the logits output by the classifier, forcibly assigning a fixed hard label to each image. This approach neglects the measurement of inter-class distances in FER, resulting in biased recognition results. To address this issue, we project the deep features into a new hidden space and then introduce a contrastive fine-tuning strategy.


Here we briefly review the concept of contrastive learning. Consider a set of $B$ sampled pairs $\{x_{k},y_{k}\}_{k=1...B}$, each image is first sent into an augmentation module and augmented as $2B$ pairs in total. $\{\widetilde{x}_{k},\widetilde{y}_{k}\}_{k=1...2B}$ are corresponding augmentations. Here we formulated $\widetilde{x}=Aug(x)$. Then $\widetilde{x}$ are encoded to the representation $r=Enc(\widetilde{x})$. After the encoder, a projection head is used to map the representation $r$ to the projection vector, $z=Proj(r)$. Then we need to select positive samples and negative samples in $2B$ pairs.

We set $B$ to the size of sample pairs in a batch. Then we call the combination of the augmented samples a multiview batch. Each multiview batch consists of $2B$ samples, one of them is chosen as the anchor, and the other $2B-1$ are candidates for positive and negative samples.
The InfoNCE loss used in contrastive learning can be formulated as:
\begin{equation}
    \mathcal{L}_{cl}=-\sum_{i=1}^{2B} log\frac {exp(sim(pos))}{ exp(sim(pos))+exp(sim(neg))},
\end{equation}
where the similarity criterion is usually based on the cosine distance so that $sim(pos)$ and $sim(neg)$ can be denoted as $(z_i \centerdot z_p)$ and $(z_i \centerdot z_n)$.
$z_i$, $z_p$, and $z_n$ represent the anchor, the positive samples, and the negative samples, respectively.
For various forms of contrastive learning, $\mathcal{L}_{cl}$ can be adjusted based on different criteria for selecting positive and negative samples.

As shown in Figure \ref{cl}, in the traditional self-supervised contrastive format, only the augmented one is seen as the positive sample, the rest of the $2B-2$ samples are considered negative samples. Thus, the self-supervised contrastive loss can be formulated as:
\begin{equation}
    \mathcal{L}_{sscl}=-\sum_{i=1}^{2B} log\frac {exp(z_i \centerdot z_{j(i)}/ \tau)}{\sum\limits_{a \in A(i)} exp(z_i \centerdot z_{a}/ \tau)}  ,
\end{equation}
where $z_{j(i)}$ represents the corresponding positive sample of anchor $z_i$, and the other $2B-1$ samples in $A(i)$ are considered the negatives. $\tau$
is a scalar temperature parameter to adjust the attention of negative samples.
Note that for self-supervised contrastive learning, there are 1 positive and $2B-2$ negative pairs for each anchor. 

\textbf{Supervised Contrastive Fine-tuning.}
However, the self-supervised contrastive format ignores intra-class similarity due to the lack of class labels and is difficult to apply in the finetuning stage. Therefore, it is necessary to introduce the supervised contrastive format. Due to the fine-grained property of FER tasks, the distinction between different categories is very limited. In fact, more than 1 sample is similar to the anchor in a multiview batch and should be considered positive samples. Therefore, $\mathcal{L}_{cl}$ can be adjusted to the supervised format:
\begin{align}
    \label{lscl}
    \mathcal{L}_{scl}&=-\sum_{i=1}^{2B}\frac{1}{|P(i)|}\sum_{p\in P(i)} log\frac {exp(z_i \centerdot z_{p}/ \tau)}{\sum\limits_{a \in A(i)} exp(z_i \centerdot z_{a}/ \tau)} ,  \\
    z_a &= Proj(Enc(\hat{x_a})) , \\
    z_{a}&=\left\{
    \begin{aligned}
    z_{p}, \hat{y_a}&=\hat{y_i} ,\\
    z_{n}, \hat{y_a}&\neq \hat{y_i} .
    \end{aligned}
    \right. 
\end{align}
Here, $P(i)\equiv \{p\in A(i): \widetilde{y_p}= \widetilde{y_i}\}$. As long as the samples in a multiview batch have the same labels as the anchor, they should be considered positive samples.

\textbf{Mix-supervised Contrastive Fine-tuning.}
Furthermore, given the extremely high inter-class similarity in the FER task (\textit{i.e.}, images from different categories may share similar visual representations that are difficult to distinguish even for human observers), we find that the samples from different categories can also provide positive supervision if they share a high similarity. Therefore, we propose to mix the image and label pairs of $2B$ augmented samples (shown in Figure\ref{cl}). 
Firstly, we sample a coefficient from the Beta distribution:
\begin{equation}
    \lambda \sim Beta(\alpha,\beta),
\end{equation}
where $\alpha$ and $\beta$ are two coefficients set to $2$ as default. Then, the samples are fused as:
\begin{equation}
    Mix(\hat{x_a},\hat{x_b}) = \lambda \hat{x_a} + (1-\lambda) \hat{x_b}.
\end{equation}
Compared to $\mathcal{L}_{scl}$, the mixed pairs provide more diverse positive samples. Even if the categories are different, samples with similar visual representations can be treated as positive samples. The samples with significant differences in both category and visual representation are treated as negative samples. Specifically, the mixture of two samples may result in one of three possible scenarios:

\textbf{Case 1} As both samples share the same label as the anchor ($\hat{y_a} = \hat{y_b}=\hat{y_i}$), their mixed label is also identical to the anchor's label ($Mix(\hat{y_a},\hat{y_b}) = \hat{y_i}$). Additionally, their visual representation after mixing is similar to that of the anchor ($r_{a,b} \approx r_i$), which leads us to select the mixed sample as a positive sample.

\textbf{Case 2} At least one of the two samples has a label that differs from that of the anchor ($\hat{y_a} \neq \hat{y_i} $ or $\hat{y_b} \neq \hat{y_i}$), but the distance between the mixed label and the anchor's label falls below a predefined threshold ($|\hat{y_i}-mix(\hat{y_a},\hat{y_b})|\leq t$), and the visual representation of the mixed sample is also similar to that of the anchor ($r_{a,b} \approx r_i$), we select the mixed sample as a positive sample.

\textbf{Case 3} At least one of the two samples has a label that differs from that of the anchor ($\hat{y_a} \neq \hat{y_i} $ or $\hat{y_b} \neq \hat{y_i}$), and the distance between the mixed label and the anchor's label exceeds a predefined threshold ($|\hat{y_i}-mix(\hat{y_a},\hat{y_b})| > t$), and the visual representation of the mixed sample is dissimilar to that of the anchor ($r_{a,b} \neq r_i$), we select the mixed sample as a negative sample.

The mixing strategy is implemented by a combination of Mixup \cite{DBLP:journals/corr/abs-1710-09412} and CutMix \cite{yun2019cutmix}. As shown in Figure \ref{fig:vis_exp}, different mixing ratio brings samples from different emotions, thereby providing more diverse positive samples for contrastive fine-tuning. The mix-supervised contrastive loss is described as:
\begin{align}
    \label{loss:mscl}
    \mathcal{L}_{mscl}&=-\sum_{i=1}^{2B}\frac{1}{|P(i)|}\sum_{p\in P(i)} log\frac {exp(z_i \centerdot z_{p}/ \tau)}{\sum\limits_{a,b \in A(i)} exp(z_i \centerdot z_{a,b}/ \tau)} ,\\ 
    z_{a,b} &= Proj(Enc(Mix(\hat{x_a},\hat{x_b})))  ,\\
    z_{a,b} &=\left\{
    \begin{aligned}
    z_{p}, |Mix(\hat{y_a},\hat{y_b})- &\hat{y_i}| \leq t, \\
    z_{n}, |Mix(\hat{y_a},\hat{y_b})- &\hat{y_i}| > t  .
    \end{aligned}
    \right. 
\end{align}

In order to force the network to learn discriminative representations, it's important to maximize the similarity among samples that share high confidence in the representation space. In this way, the knowledge learned on image classification tasks can be transferred well to FER tasks. We perform end-to-end contrastive fine-tuning by combining the classification loss and the mix-supervised contrastive loss with $\lambda$ as a trade-off parameter. The total loss function in the fine-tuning stage is:
\begin{equation}
    \mathcal{L}_{all}^{f}=\mathcal{L}_{cls}^{f} + \lambda \mathcal{L}_{mscl}^{f} .
    \label{ltotal}
\end{equation}

\begin{algorithm}[t]
\caption{Training procedure of MIMIC}
\label{alg}
\textbf{Input}: General data $D_{g}$ and FER data $D_{f}$\\
\textbf{Output}: Network $\theta_n^{(T)}$\\
\textbf{Initialize}: $\theta_n^{(0)}$
\begin{algorithmic}[1] 
\FOR{t = 0...$T^{'}-1$}
\STATE $x_{g}^{i}$  $\gets$ SampleMiniBatch ($D_{g}$) \\
\STATE $x_{g}^{m}$  $\gets$ RandomMask ($x_{g}^{i}$) \\
\STATE $x_{g}^{o}$  $\gets$ Forward ($x_{g}^{m}$, $\theta_{Enc}^{(t)}+\theta_{Dec}^{(t)} $) \\
\STATE Update the Enocder and the Decoder using Eq. \ref{lrec}.\\
\ENDFOR
\FOR{t = $T^{'}$...$T-1$}
\STATE $x_{f}$, $y_{f}$ $\gets$ SampleMiniBatch ($D_{f}$) \\
\STATE $\widetilde{x}_{f}$, $\widetilde{y}_{f}$ $\gets$ Aug ($x_{f}$, $y_{f}$) \\
\STATE $r_{f}$  $\gets$ Forward ($Mix(\widetilde{x}_{f})$, $\theta_{Enc}^{(t)}$) \\
\STATE $z_{f}$  $\gets$ Forward (${r}_{f}$, $\theta_{Proj}^{(t)}$) \\
\STATE $\hat{y}_{f}$  $\gets$ Forward (${r}_{f}$, $\theta_{Cls}^{(t)}$) \\
\STATE Simultaneously distinguish between positive and negative samples and update the classification head using Eq. \ref{ltotal}.
\ENDFOR
\STATE \textbf{return} $\theta_N^{(T)}$\\
\end{algorithmic}
\end{algorithm}

\subsection{Algorithm for MIMIC}
We list step-by-step pseudo-code for our method in Algorithm \ref{alg}. The first stage from line 1 to 6 is the mask image pre-training on general image classification datasets. We first randomly mask the input images and update the network using Equation \ref{lrec}. After pre-training, the encoder is taken for contrastive fine-tuning on FER datasets from line 7 to 14, while the decoder is dropped. In the second stage, we simultaneously optimize the distance between representations by applying Equation \ref{loss:mscl} and update the classification head leveraging Equation \ref{lfcls}. Note that $r_{f}$ is used to update the classification head while $z_{f}$ is utilized to aggregate positive samples through a projection head. After fine-tuning, the network learns good facial expression representations from general images and generalizes well to the FER task.

\section{Experiment}
\label{expriment}

\subsection{Datasets}
{\textbf{RAF-DB} \cite{li2017reliable} comprises around 30000 facial images that have been annotated by 40 trained human coders with basic or compound expressions. For our study, we utilize images that display the seven basic expressions (neutral, happiness, surprise, sadness, anger, disgust, and fear). This results in a total of 12,271 images for training and 3,068 images for testing. 

\textbf{FERPlus} \cite{barsoum2016training}, an extension of FER2013 used in the ICML 2013 Challenges, is collected by the Google search engine. It contains 28,709 training images, 3,589 validation images, and 3,589 test images. Unlike FER2013, FERPlus includes contempt as an additional emotion category, resulting in a total of 8 classes. 

\textbf{AffectNet} \cite{mollahosseini2017affectnet}, the largest dataset to provide both categorical and Valence-Arousal annotations, contains over one million images obtained from the Internet by querying expression-related keywords in three search engines. Of these images, 450,000 have been manually annotated with eight expression labels, as in FERPlus. The dataset includes imbalanced training and test sets, as well as a balanced validation set. According to previous work, there are two different divisions of AffectNet7 and AffectNet8. We select AffectNet7 to evaluate performance, which lacks the expression of contempt with 3667 training images and 500 test images.

\subsection{Implmentation Details}
For a fair comparison, we select ViT-base/16 as the backbone and ImageNet-1k\cite{imagenet} as the pre-training dataset to evaluate the performance. In the pre-training stage, the mask strategy and training epochs are set based on the analysis in \cite{he2022masked}. The threshold $t$ is pre-defined at 0.5. The other hyperparameters are set according to the analysis of parameter sensitivity. All the images are aligned and resized to $224\times224$. We augment two cropped patches per image following the augmentation strategy in \cite{xie2022simmim}. We use Adam optimizer for the three FER datasets. The weight decay is 5e-4 and the learning rate is initialized as 1e-4 with 0.65 layer decay. For AffectNet, a balanced sample strategy is taken to reduce the influence of unbalanced training set distribution. We fine-tune RAF-DB and FERPlus for 50 epochs, while AffectNet for 10 epochs. All the experiments are implemented in Pytorch with NVIDIA A100 GPUs.

\begin{table*}[th]
\centering
\caption{Comparison with the state-of-the-art methods. Overall Accuracy ($\%$) is reported on the test dataset.}
\label{sota}
\resizebox{0.9\textwidth}{!}{
\begin{tabular}{l|ccccc}
\toprule
Method & Backbone& Pre-train Data & RAF-DB  & FERPlus  & AffectNet \\
\midrule
DDA \cite{farzaneh2020discriminant}        & ResNet-18          & MS-Celeb-1M   & 86.90                        & -                            & 62.34      \\
SCN \cite{wang2020suppressing}       & ResNet-18          & MS-Celeb-1M   & 87.03                        & 88.01                        & 63.40      \\
SCAN \cite{gera2021landmark}      & ResNet-50          & MS-Celeb-1M   & 89.02                        & 89.42                        & 65.14      \\
DMUE \cite{she2021dive}      & ResNet-18          & MS-Celeb-1M   & 88.76                        & 88.64                        & -          \\
RUL \cite{zhang2021relative}       & ResNet-18          & MS-Celeb-1M   & 88.98                        & 88.75                        & 61.43      \\
DACL \cite{farzaneh2021facial}      & ResNet-18          & MS-Celeb-1M   & 87.78                        & 88.39                        & 65.20      \\
MA-Net \cite{zhao2021learning}     & ResNet-18          & MS-Celeb-1M   & 88.40                        & -                            & 64.53      \\
MVT \cite{li2021mvt}       & DeIT-S/16          & ImageNet-1K    & 88.62                        & 89.22                        & 64.57      \\
VTFF \cite{ma2021facial}       & ResNet-18+ViT-B/32          & MS-Celeb-1M   & 88.14                        & 88.81                        & 64.80      \\
EAC \cite{zhang2022learn}        & ResNet-18          & MS-Celeb-1M   & 89.99                        & 89.64                        & 65.32      \\
RC \cite{feng2020provably}        & ResNet-18          & MS-Celeb-1M   & 88.92                        & 88.53                        & 64.87      \\
LW \cite{wen2021leveraged}        & ResNet-18          & MS-Celeb-1M   & 88.70                        & 88.10                        & 64.53      \\
PICO \cite{wang2022pico}      & ResNet-18          & MS-Celeb-1M   & 88.53                        & 87.93                        & 64.35      \\
APVIT \cite{xue2022vision}      & ResNet-50          & MS-Celeb-1M   & 90.53                        & 89.35                        & 63.92      \\
\midrule
MIMIC (Ours)       & ViT-B/16           & ImageNet-1K    & 89.02                        & 89.74                        &     65.35       \\
MIMIC (Ours)       & ViT-L/16           & ImageNet-1K    & \textbf{91.26} & \textbf{91.14} &  \textbf{65.62}\\
\bottomrule
\end{tabular}%
}
\end{table*}
\begin{figure*}[th]
    \centering
    \includegraphics[width=0.9\linewidth]{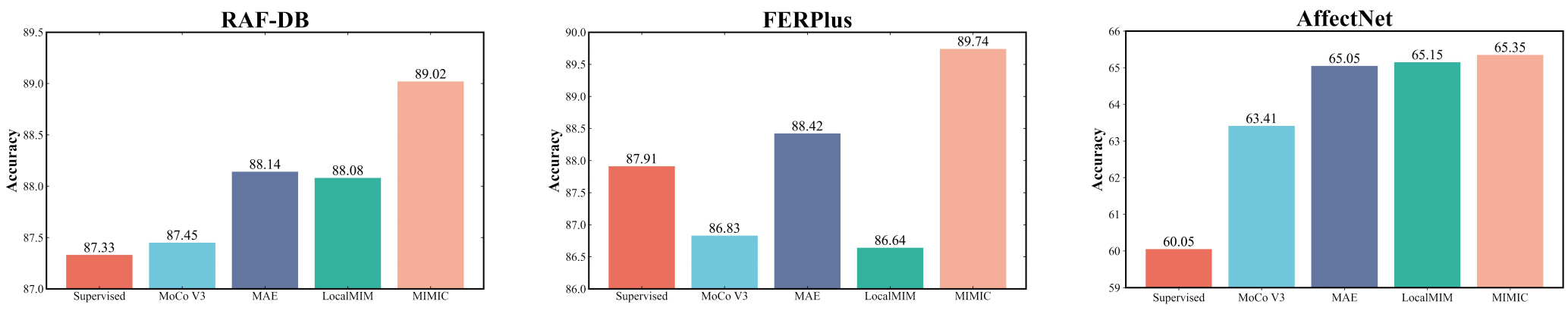}
    \caption{Comparison with other pre-training methods. All the methods are pre-trained on ImageNet-1k and then fine-tuned on FER datasets. Our method outperforms supervised, MIM (\textit{e.g.}, MAE, LocalMIM), and contrastive learning (\textit{e.g.}, MoCo V3) methods.}
    \label{exp}
\end{figure*}

\subsection{Comparison with Baselines}
We select four methods from three different training strategies as our baselines. All the methods take ViT-B/16 as the backbone and a simple classification head to output the prediction. These baselines include the fully supervised method, contrastive learning method (MoCo V3 \cite{chen2021mocov3}), and MIM methods (MAE \cite{he2022masked} and LocalMIM \cite{wang2023masked}). Results on the three datasets in Figure \ref{exp} show that our method outperforms all the baselines. On RAF-DB and AffectNet, both contrastive learning and MIM methods outperform the traditional supervised pre-training strategy. Our method leverages the advance of strong learning representation from MIM and bridges the gap between general images and FER datasets by contrastive learning. Therefore, our method achieves higher performance on these two datasets. On FERPlus, LocalMIM and MoCo V3 are not superior to the supervised pre-training strategy. This phenomenon reflects the domain disparity between FERPlus and ImageNet is considerable and naive fine-tuning cannot bridge this gap. However, by leveraging contrastive fine-tuning, our method easily outperforms the supervised method and other unsupervised methods, which shows the strong ability to correct the similarity of the representations in the latent space.

\begin{table*}[t]
\centering
\caption{Ablation on the proposed components about pre-training and fine-tuning stage. SCL and MSCL denote supervised contrastive loss and mix-supervised contrastive loss, respectively.}
\label{ablation}
\resizebox{0.8\linewidth}{!}{%
\begin{tabular}{cc|ccc|lll}
\toprule
\multicolumn{2}{c|}{Pre-training} & \multicolumn{3}{c|}{Fine-tuning} & \multirow{2}{*}{RAF-DB} & \multirow{2}{*}{FERPlus} & \multirow{2}{*}{AffectNet} \\
Supervised & MIM & CE & SCL & MSCL &       &       &       \\
\midrule
\checkmark          &     & \checkmark  &     &      & 87.33 & 87.91 & 60.05 \\
\checkmark          &     & \checkmark  & \checkmark   &      & 87.52 \textcolor{black}{$\uparrow$ 0.19} & 88.06 \textcolor{black}{$\uparrow$ 0.15} & 63.10 \textcolor{black}{$\uparrow$ 3.05}\\
\checkmark          &     & \checkmark  &     & \checkmark    & 87.60 \textcolor{black}{$\uparrow$ 0.27} & 88.21 \textcolor{black}{$\uparrow$ 0.30} & 63.62 \textcolor{black}{$\uparrow$ 3.57} \\
\midrule
           & \checkmark   & \checkmark  &     &      & 88.14 \textcolor{black}{$\uparrow$ 0.81} & 88.42 \textcolor{black}{$\uparrow$ 0.51} & 65.05 \textcolor{black}{$\uparrow$ 5.00} \\
           & \checkmark   & \checkmark  & \checkmark   &      & 88.87 \textcolor{black}{$\uparrow$ 1.54} & 89.43 \textcolor{black}{$\uparrow$ 1.52} & 65.26 \textcolor{black}{$\uparrow$ 5.21} \\
           & \checkmark   & \checkmark  &     & \checkmark    & 89.02 \textcolor{black}{$\uparrow$ 1.69} & 89.74 \textcolor{black}{$\uparrow$ 1.83} & 65.35 \textcolor{black}{$\uparrow$ 5.30}\\
\bottomrule
\end{tabular}%
}
\end{table*}

\begin{table}[t]
\centering
\caption{Ablation on projection head with ViT-B/16 as the backbone.}
\label{proj_head}
\resizebox{\columnwidth}{!}{%
\begin{tabular}{l|lll}
\toprule
Projection Head & RAF-DB          & FERPlus & AffectNet       \\
\midrule
None            & 87.68          & 88.65     & 65.14     \\
Linear          & 87.18 \textcolor{black}{$\downarrow$ 0.50}          & 88.43 \textcolor{black}{$\downarrow$ 0.22} & 64.62 \textcolor{black}{$\downarrow$ 0.52}        \\
Dense           & 89.02 \textcolor{black}{$\uparrow$ 1.34} & 89.74 \textcolor{black}{$\uparrow$ 1.09} & 65.35 \textcolor{black}{$\uparrow$ 0.21}\\
\bottomrule
\end{tabular}%
}
\end{table}
\vspace{-3mm}
\begin{table}[t]
\centering
\caption{Ablation on classification head with ViT-B/16 as the backbone.}
\label{cls_head}
\resizebox{\columnwidth}{!}{%
\begin{tabular}{l|lll}
\toprule
Classification Head & RAF-DB          & FERPlus  & AffectNet      \\
\midrule
Class Token         & 86.38          & 88.08 & 63.67         \\
Global Average Pooling                 & 89.02 \textcolor{black}{$\uparrow$ 2.64} & 89.74 \textcolor{black}{$\uparrow$ 1.66} & 65.35 \textcolor{black}{$\uparrow$ 1.68}\\
\bottomrule
\end{tabular}%
}
\end{table}
\vspace{-3mm}
\begin{table}[t]
    \centering
    \caption{FER accuracy comparison ($\%$) with respect to different pre-training domains using ViT/B-16 as the backbone.}
    \label{tab: pretrain-domain}
    \resizebox{\columnwidth}{!}{%
    \begin{tabular}{l|lll}
    \toprule
        Pre-training & RAF-DB & FERPlus & AffectNet \\
    \midrule
        MS-Celeb-1M & 86.18 & 87.35 & 61.62 \\
        ImageNet-1K & 89.02 \textcolor{black}{$\uparrow$ 2.84}& 89.74 \textcolor{black}{$\uparrow$ 2.39}& 65.35 \textcolor{black}{$\uparrow$ 3.73}\\
    \bottomrule
    \end{tabular}%
    }
\end{table}

\subsection{Comparison with State-of-the-Art}
We show that by scaling up the model size (\textit{i.e.}, from ViT-B/16 to ViT-L/16), the performance of our approach is not saturated. Results in Table \ref{sota} show that compared with the state-of-the-art (SOTA) methods, our approach still achieves impressive results by taking ViT-L/16 as the backbone. Previous methods usually take ResNet-18 or 50 as the backbone, for larger model structures do not necessarily enhance the performance. They usually design some complex networks after the backbone to stimulate performance improvements. A common understanding in FER is that ViT pre-trained on ImageNet as a backbone performs worse than ResNet pre-trained on MS-Celeb-1M. Our method challenges this notion and breaks people's perception of this idea. We demonstrate that without modifying the architecture, it is possible to achieve SOTA performance with vanilla ViT by simply employing advanced training strategies.

\subsection{Ablation Study}
\textbf{Effectiveness of Components.}
To verify the effectiveness, we select ViT-B/16 with pre-trained weights on ImageNet-1k as our baseline. The hyperparameter setting of MIM follows the default setting in MAE \cite{he2022masked}. 
From the results, the following conclusions can be observed:
Firstly, compared to supervised pre-training, mask image pre-training achieves performance improvements on the three FER datasets. When fixing the fine-tuning manner, comparisons between the first row and the fourth row, the second row and the fifth row, and the third row and the sixth row demonstrate that mask image pre-training can outperform the previous supervised pre-training method with no need to label supervision.
Secondly, experimental results demonstrate that under the supervised pre-training setting and the mask image pre-training setting, both SCL and MSCL enhance the performance. Our proposed contrastive fine-tuning strategy with MSCL is superior to both the conventional fine-tuning strategy with CE loss and the fine-tuning strategy with SCL.
By combining the two stages (the last row) together, our approach achieves SOTA performance in the ViT architecture.

\textbf{Ablation on Projection Head.}
Which latent space is more suitable for fine-tuning, before or after the projection head? This is a frequently discussed problem in the field of contrastive learning. Generally, it is often believed that there is a stronger and more stable representational capacity in the space after the projection head. The commonly used projection heads can be divided into two types: the linear projection head composed of a single linear layer, and the dense projection head composed of linear-ReLU-linear layers. In our experiments on three datasets shown in Table \ref{proj_head}, we test the effects of different projection heads. The experimental results show that the linear projection head performs the worst, while the dense projection head shows a stable improvement compared to not using a projection head.

\textbf{Ablation on Classification Head.}
In ViT, different classification heads usually do not bring about differences in classification results. However, when the data in the pre-training and fine-tuning stages come from different domains, the experimental results in Table \ref{cls_head} show that using global average pooling (GAP) often brings better results than using the class token. GAP focuses on the average information of the data, while class token focuses on the most important information. In the fine-tuning stage, the core is to narrow the gap between the two domains, so concentrating on the average information is more appropriate than only focusing on the most important information.

\textbf{Ablation on the Pre-training Domain.}
Due to the presence of relatively similar facial images between the FER dataset and the face dataset, the results of MIM pre-training on a large-scale face dataset are also of interest.
However, results shown in Table \ref{tab: pretrain-domain} demonstrate that pre-training on the face recognition dataset MS-Celeb-1M does not yield better outcomes than pre-training on the general image dataset ImageNet-1K. We deduce this is attributed to the fact that images in face recognition datasets are not diverse enough for MIM, making the reconstruction task too singular and simple, thereby allowing the network to learn trivial solutions. Therefore, we prefer to fix the pipeline as pre-training on ImageNet-1K and then fine-tuning on FER datasets.

\begin{figure}[t]
    \centering
    \includegraphics[width=\linewidth]{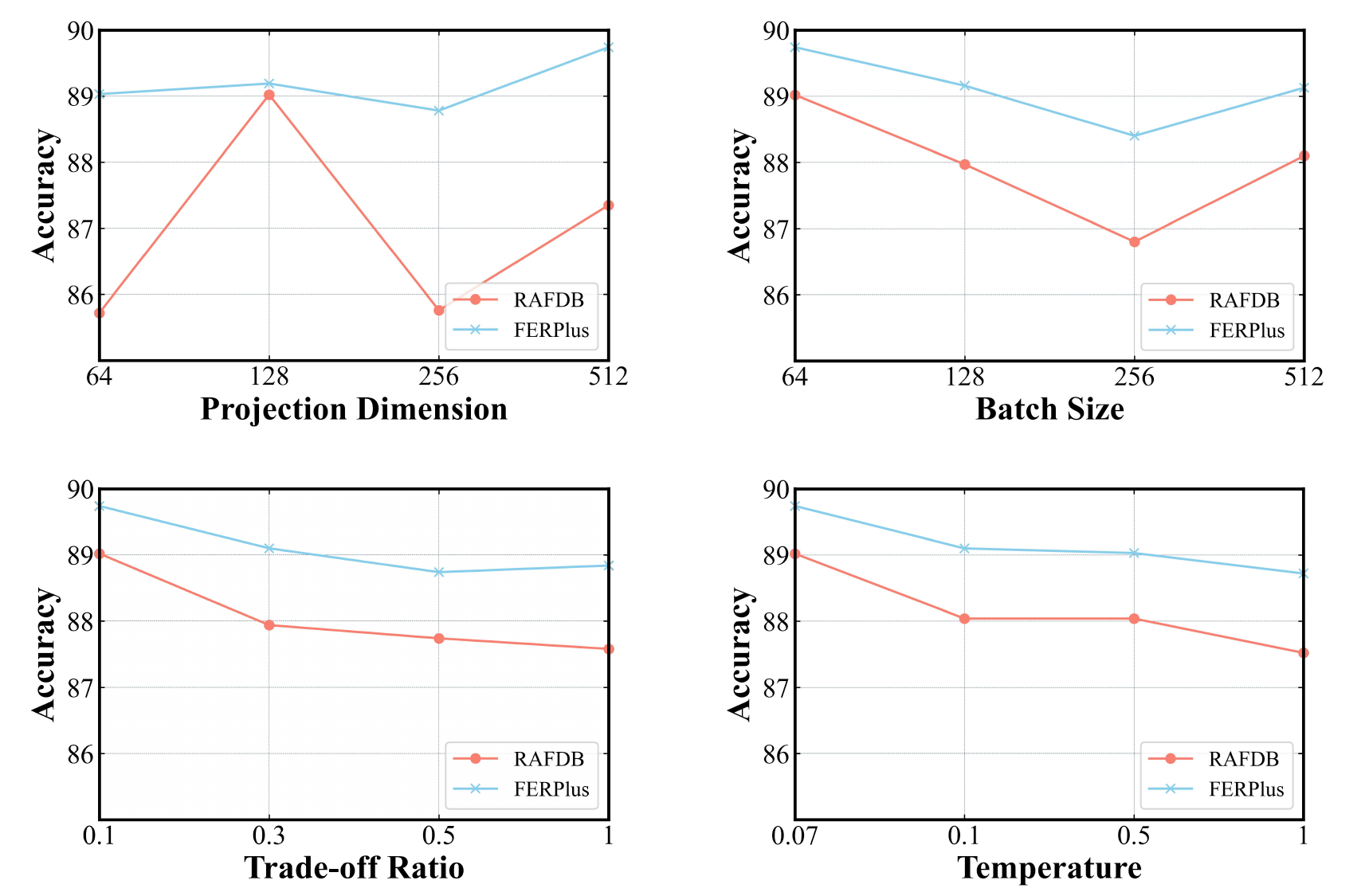}
    \caption{Parameter sensitivity on projection dimension, batch size, trade-off ratio $\lambda$, and temperature $\tau$.}
    \label{parameter}
\end{figure}

\subsection{Parameter Sensitivity}
Figure \ref{parameter} shows the performance of our method with different projection dimension, batch size, trade-off ratio $\lambda$, and temperature $\tau$ on two datasets. 

On RAF-DB, a dimension of 128 performs best, while a dimension of 512 performs best on FERPlus. Batch size refers to the number of positive samples and negative samples. In self-supervised contrastive learning, larger batch size is generally better because it provides more negative samples. However, in FER, the mix-supervised contrastive fine-tuning is different. Since label information is utilized, the positive and negative samples are more accurate, so a larger batch size is unnecessary to provide more negative samples. In addition, the number of categories in FER is particularly small (usually only 7 or 8), an excessively large batch size may lead to overfitting. In our experiments, results from both datasets indicate that a batch size of 64 exhibits the best performance. The trade-off ratio $\lambda$ aims to improve the representation capability while reducing the gap between two domains when training a classifier. In our experiments, a trade-off ratio of 0.1 exhibits the best performance. The temperature $\tau$ is used to increase the penalty for negative samples, and is typically set to a value less than 1, with a classical setting of 0.07. Our test results on two datasets indicate that, for the FER task, the optimal value of the temperature $\tau$ is the same as that for traditional image classification tasks.

\begin{figure}[t]
    \centering 
    \includegraphics[width=\linewidth]{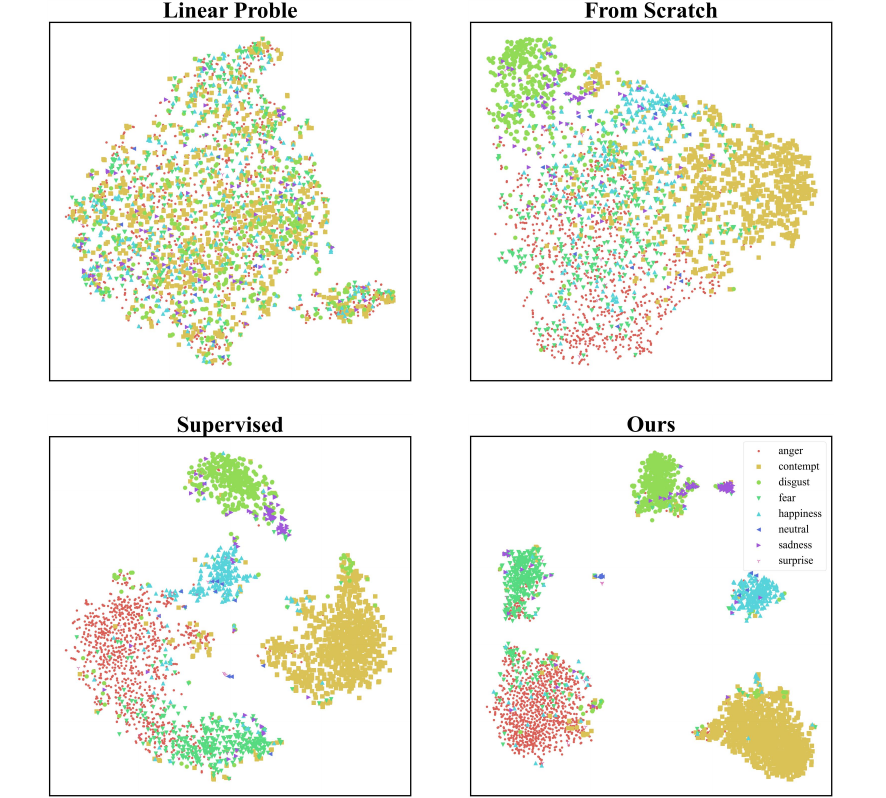}
    \caption{t-SNE visualizations in a global view on FERPlus validation set. The embeddings are learned by different methods. All the methods employ the ViT-B/16 network architecture, with only variations in their training strategies.}
    \label{tsne}
\end{figure}

\subsection{Qualitative Analysis}
We make the 2-D visualization (t-SNE \cite{van2008visualizing}) for the feature embeddings from the last layer of the ViT-B/16 encoder learned by four different methods in Figure \ref{tsne}. We first visualize the features obtained from linear probing and training from scratch. Linear probing refers to freezing the encoder pre-trained by MIM and only training the classifier. Training from scratch represents discarding pre-training from other domains, but rather conducting end-to-end training with randomly initialized parameters. The two methods in the first row lack the fine-tuning and pre-training stages respectively, resulting in representations corrupted by adversaries and demonstrating poor class discrimination. Proceeding, we employ the same fine-tuning techniques to compare supervised pre-training with our mask image pre-training approach. The two methods in the second row are pre-trained in different ways, and it can be observed that our method yields better class separation compared to these baselines. This demonstrates that our method effectively mitigates instance-level identity confusion.

\section{Conclusion}
In this paper, we divide the methods to solve the FER task into two stages: pre-training and fine-tuning, and make improvements in both stages compared to the conventional DFER paradigm. In the pre-training stage, we replace supervised pre-training on the large-scale face dataset MS-Celeb-1M with self-supervised mask image pre-training on the mid-scale general dataset ImageNet-1K. This improvement reduces the expense of pre-training and mitigates the cost of label annotation.
In the fine-tuning stage, we introduce a mix-supervised contrastive branch to the traditional fine-tuning methods to reduce the gap between FER and other domains. 
Through extensive experiments on several benchmark datasets, we also demonstrate that it is possible to achieve SOTA performance by leveraging vanilla ViT and advanced training strategies.
In future research endeavors, we aim to broaden the scope of our methodology to encompass other intricate domains of affective computing, \textit{e.g.}, multimodal emotion recognition, and video-based emotion analysis.

\section*{Acknowledgement}
The authors are grateful to the anonymous reviewers for critically reading the manuscript and for giving important suggestions to improve their paper.

\bibliographystyle{IEEEtran}
\bibliography{rec}

\begin{IEEEbiography}
[{\includegraphics[width=1in,height=1.25in]{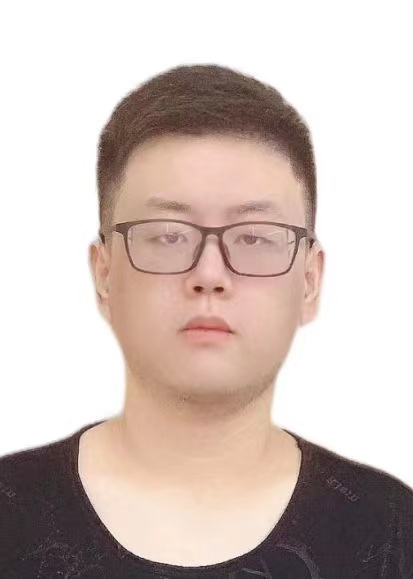}}]
{Fan Zhang} is working towards his M.S. degree in Electrical and Computer Engineering, at Shenzhen campus, Georgia Institute of Technology. He received the B.S. degree in Mechatronic Engineering from Soochow University, Suzhou, China in 2022. His research interests include affective computing, computer vision, and deep learning.
\end{IEEEbiography}
\vspace{-10mm}
\begin{IEEEbiography}
[{\includegraphics[width=1in,height=1.25in,clip,keepaspectratio]{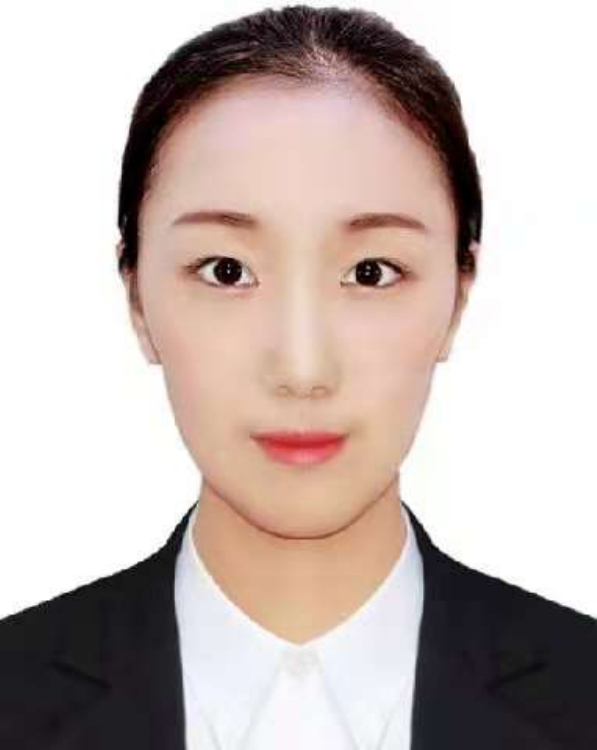}}]{Xiaobao Guo} Xiaobao Guo is a Ph.D. student with the School of Computer Science and Engineering, and with Rapid-Rich Object Search (ROSE) Lab, Nanyang Technological University. Her research interests include computer vision and multimodal learning. 
\end{IEEEbiography}
\vspace{-10mm}
\begin{IEEEbiography}
[{\includegraphics[width=1in,height=1.25in]{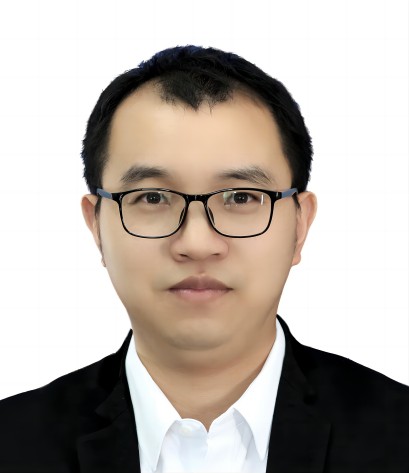}}]
Xiaojiang Peng is a full professor at the College of Big Data and Internet, Shenzhen Technology University. He received his Ph.D. degrees from Southwest Jiaotong University. He was an associate professor at Shenzhen Institutes of Advanced Technology Chinese Academy of Sciences from 9/10/2017 to 31/10/2020. He was a postdoctoral researcher at Idiap, Switzerland from 1/7/2016 to 30/9/2017 where he worked with Prof. Francois Fleuret. He was a postdoctoral researcher at Inria LEAR/THOTH, France from 1/3/2015 to 30/6/2016 where he worked with Cordelia Schmid. He has published more than 70 top journal/conference papers (e.g., TIP, CVPR, ICCV, ECCV, IJCAI, AAAI). His research interests include computer vision, affective computing, and deep learning.
\end{IEEEbiography}
\vspace{-10mm}
\begin{IEEEbiography}
[{\includegraphics[width=1in,height=1.25in,clip,keepaspectratio]{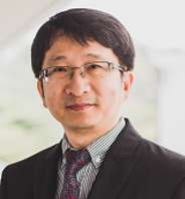}}]{Alex Kot}Prof. Alex Kot has been with the Nanyang Technological University, Singapore since 1991. He was Head of the Division of Information Engineering and Vice Dean Research at the School of Electrical and Electronic Engineering. Subsequently, he served as Associate Dean for College of Engineering for eight years. He is currently Professor and Director of Rapid-Rich Object SEarch (ROSE) Lab and NTU-PKU Joint Research Institute. He has published extensively in the areas of signal processing, biometrics, image forensics and security, and computer vision and machine learning.  

Dr. Kot served as Associate Editor for more than ten journals, mostly for IEEE transactions. He served the IEEE SP Society in various capacities such as the General Co-Chair for the 2004 IEEE International Conference on Image Processing and the Vice-President for the IEEE Signal Processing Society. He received the Best Teacher of the Year Award and is a co-author for several Best Paper Awards including ICPR, IEEE WIFS and IWDW, CVPR Precognition Workshop and VCIP. He was elected as the IEEE Distinguished Lecturer for the Signal Processing Society and the Circuits and Systems Society. He is a Fellow of IEEE, and a Fellow of Academy of Engineering, Singapore.
\end{IEEEbiography}

\end{document}